\documentclass[10pt, a4paper]{article}

\usepackage[final]{lrec-coling2024} 
\usepackage{booktabs}
\usepackage{tabularx}
\usepackage{adjustbox}
\usepackage{graphicx}
\usepackage{color}
\usepackage{float}

\title{Chitchat as Interference: \\
Adding User Backstories to Task-Oriented Dialogues}

\name{Armand Stricker, Patrick Paroubek} 

\address{ Université Paris-Saclay, CNRS, Laboratoire Interdisciplinaire des Sciences du Numérique\\
         91400, Orsay, France  \\
         \{armand.stricker, patrick.paroubek\}@lisn.upsaclay.fr\\}

\abstract{
During task-oriented dialogues (TODs), human users naturally introduce chitchat that is beyond the immediate scope of the task, interfering with the flow of the conversation. To address this issue without the need for expensive manual data creation, we use few-shot prompting with Llama-2-70B to enhance the MultiWOZ dataset with user backstories,  a typical example of chitchat interference in TODs. We assess the impact of this addition by testing two models: one trained solely on TODs and another trained on TODs with a preliminary chitchat interaction. Our analysis demonstrates that our enhanced dataset poses a challenge for these systems. Moreover, we demonstrate that our dataset can be effectively used for training purposes, enabling a system to consistently acknowledge the user's backstory while also successfully moving the task forward in the same turn, as confirmed by human evaluation.  These findings highlight the benefits of generating novel chitchat-TOD scenarios to test TOD systems more thoroughly and improve their resilience to natural user interferences.\\ \newline  }

\begin{document}

\maketitleabstract

\section{Introduction}

Chitchat and task-oriented dialogue (TOD) agents are typically portrayed as two distinct systems. On the one hand, chitchat agents are expected to embody all the qualities of an ideal conversationalist. They should be empathetic, engaging, knowledgeable, and well-behaved \cite{roller2020opendomain, smith-etal-2020-put, rashkin-etal-2019-towards}. On the other hand, task agents are designed to be efficient and effective tools \cite{chen2017survey, deriu2021survey}. 

However, natural human communication does not make such a clear-cut distinction: most language is not purely transactional\footnote{the focus of the encounter is external and leads to an action} or interactional\footnote{the focus of the encounter is internal, and is centered on the relationship between participants} but rather a mix of both \cite{brown1983teaching}, and overlap is therefore common in real-world interactions. 

\begin{figure}[!b]
\centering
\includegraphics[width=7.5cm]{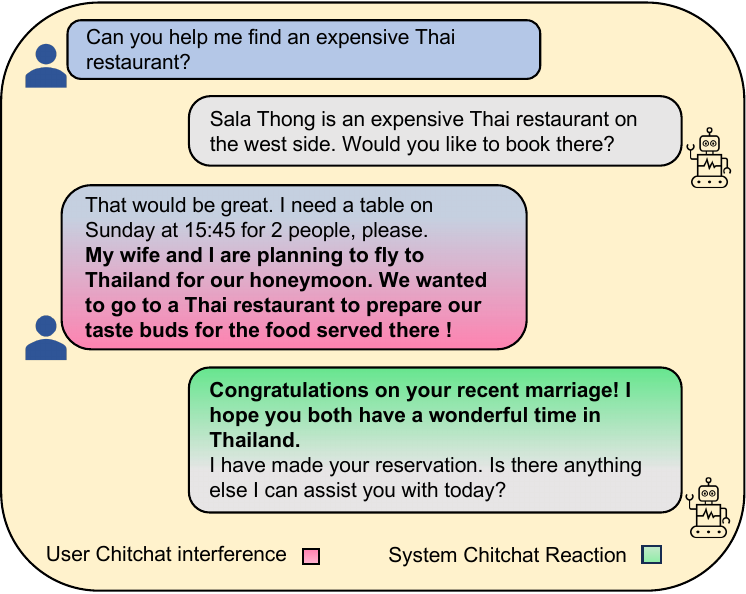}
\caption{\small A chatty user incorporates elements of backstory to their task-oriented request, creating a natural interference in the TOD interaction. The system reaction accommodates the additional backstory with support and understanding, all the while avoiding the introduction of new topics. This design choice ensures that the system seamlessly transitions back to the task at hand, effectively assisting the user in achieving their goal.}
\label{fig:Interference}
\end{figure}

Efforts have been undertaken to augment TOD datasets  with chitchat, in order to move towards more flexible TODs \cite{li-etal-2023-enhancing-task, young2022fusing}.  These augmentations are automatic as well as human-generated and produce more interesting dialogues \cite{2023strickerenhancing}, which humans tend to prefer. However, it is important to note that each individual turn in these datasets is categorized as being strictly chitchat or task-oriented. As a result, it remains unclear how a system trained on such data might react to user turns in which chitchat and task talk are seamlessly combined (Figure \ref{fig:Interference}).

These inter-mode user turns are important to consider as they can commonly occur in human-chatbot conversations. Indeed, in their analysis of live customer service chat logs, \citealp{beaver_towards_2020} noted that users often engage in self-disclosure when making requests. For example, in the travel domain, users frequently emphasize the significance of their trips by sharing details about the reasons of their travel or the people they are visiting. In this work, we examine this form of interference more closely.

Although \citealp{beaver_towards_2020} have made a portion of the studied data accessible, it only includes the initial user turns from each conversation. This limitation, resulting from privacy restrictions related to live company chat logs, complicates the task of evaluating and potentially enhancing TOD agents in similar situations for researchers.

To remedy this gap, we present a dataset that includes instances where a chatty user introduces elements of backstory following their task-oriented request. In its response, the system first reacts to the user backstory  before continuing with the task. Such user turns have the potential of derailing the TOD interaction, and thereby challenge the system to skillfully manage both the chitchat and the redirection of the conversation back to the task (Figure \ref{fig:Interference}).

Crafting these scenarios manually is strenuous. We consequently set out to create these examples automatically, deriving them from human-generated chitchat exchanges in FusedChat \cite{young2022fusing} (Section \ref{sec:approach}), a version of MultiWOZ \cite{budzianowski-etal-2018-MultiWOZ} augmented with full chitchat exchanges. By leveraging the information presented by the user in these exchanges, we automatically enhance a number of MultiWOZ dialogues. 

To evaluate the effects of the added chitchat interferences, we utilize SimpleToD \cite{hosseini-asl_simple_2020}, a popular end-to-end approach that relies on a single language model (Section \ref{sec:method}). Given the impressive performances of recent state-of-the-art models, we create a strong baseline by combining this approach with one such model, LLama-2-7B \cite{touvron2023llama}, and LoRA \cite{hu2021lora}, a parameter-efficient fine-tuning technique.

We use this approach to train two baseline systems. For a system trained solely on TOD, we adhere to the standard training protocol for \textbf{SimpleToD}, utilizing unaugmented MultiWOZ dialogues. To test a system that has been exposed to chitchat under a different form than in our enhancements, we train a \textbf{SimpleToD-fused} variant on FusedChat dialogues. These are MultiWOZ dialogues to which a chitchat exchange has been prepended (Appendix \ref{appendix:fusedchat_example}). Furthermore, to estimate the effectiveness of our automatically-generated data for training purposes, we train a \textbf{SimpleToD-inter} version. We find our data can indeed allow a model to smoothly and consistently handle user backstories while advancing the task in the same turn.

Overall, we make the following contributions:
\begin{itemize}
    \item We introduce a novel task that treats user chitchat as a disruptive element in TODs, merging task-oriented requests and elements of backstory into a single user turn.
    \item We automatically construct a dataset which may serve as a testbed for TOD systems. 
    \item We carry out experiments to assess the level of difficulty posed by our task for the popular SimpleToD baseline and its variant SimpleToD-fused, trained with a state-of-the-art LLM. We outline the shortcomings and successes of each variant, drawing from both automatic and human evaluations (Sec. \ref{sec: results}).
    \item We show a system trained on our automatically generated data (SimpleToD-inter) can effectively  integrate chitchat reactions into its responses, moving the task forward while making the user feel heard.
\end{itemize}

Our experimental code and generated data can be found on GitHub\footnote{\url{https://github.com/armandstrickernlp/chitchat-as-interference}}.

\section{Proposed Augmentation Pipeline} \label{sec:approach}

To augment the well-known MultiWOZ benchmark \cite{budzianowski-etal-2018-MultiWOZ}, we propose to re-purpose elements from the FusedChat dataset \cite{young2022fusing}, also a MultiWOZ extension.  FusedChat both prepends and appends human-generated chitchat exchanges to the pre-existing TODs. We leverage the \textit{prepended} exchanges to simulate our scenario, as these tend to portray a plausible context for the user's interaction, valuable for generating backstories (Appendix \ref{appendix:fusedchat_example}).

\begin{figure*}[ht]
\centering
\includegraphics[width=16cm]{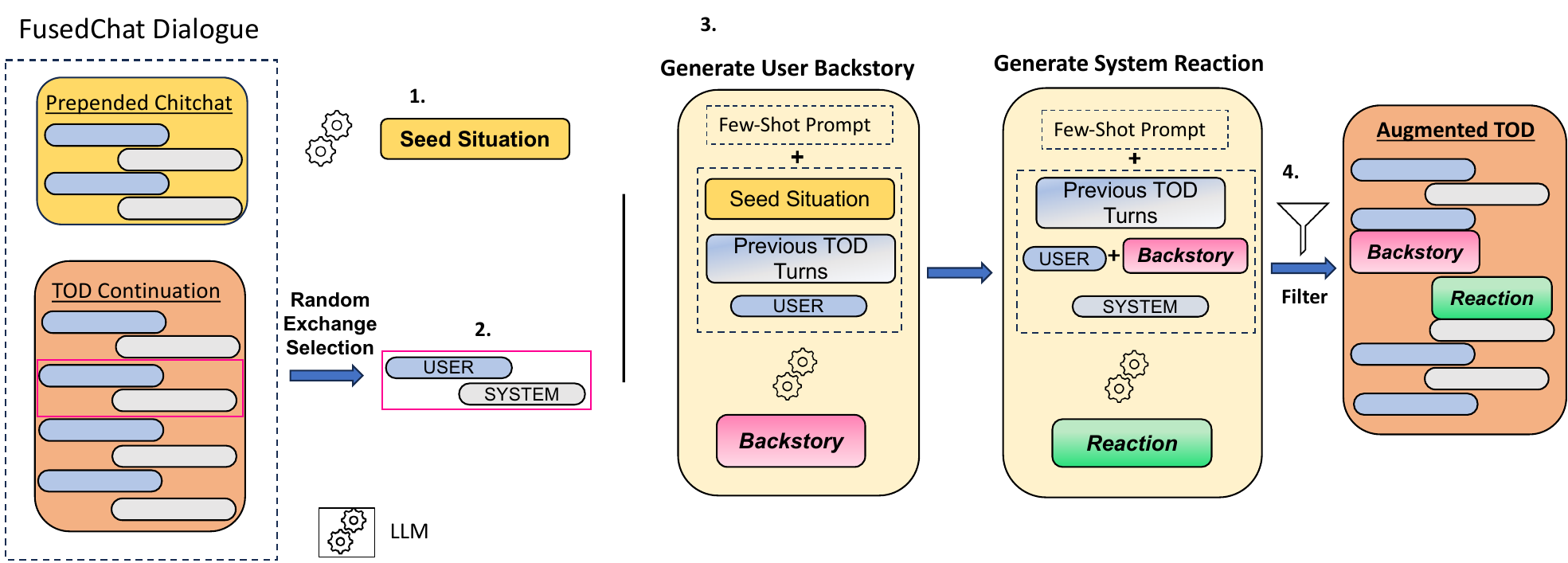}
\caption{\small Our proposed augmentation generation pipeline. Given a Fusedchat dialogue, we 1) summarize the prepended chitchat into a seed situation, 2) select a random exchange to augment from the task continuation, 3) expand the exchange with  in-context learning by first generating the backstory and then the chitchat reaction, 4) filter out potentially low quality generations.}
\label{fig:Pipeline}
\end{figure*}

We reduce the need for human involvement by automating the creation of backstories and chitchat reactions to add to the user and system turns, respectively. Leveraging fast and automated data generation methods has become a promising avenue in dialogue research, given the in-context learning capabilities of recent LLMs. Comparable approaches have recently been employed to create annotated TODs  \cite{li-etal-2022-controllable}, as well as chitchat dialogues grounded in personas \cite{lee-etal-2022-personachatgen}. We note that compared with generating entire conversations from scratch, generating our augmentations is a more manageable process less likely to yield incoherent exchanges. Indeed, we keep the original task dialogue intact and the backstories we generate are based on contextual exchanges.

We leverage Llama-2-70B-chat\footnote{\url{https://huggingface.co/meta-llama/Llama-2-70b-chat-hf}}, an open-source LLM trained on 2 trillion tokens of data from publicly accessible sources \cite{touvron2023llama}. It has additionally been fine-tuned by the authors for chat. In our preliminary experiments, this particular model demonstrated higher effectiveness in instruction-following. It also generated friendlier and more diverse system reactions compared with other Llama-2 models. Hence, we opt for this model to maximize the quality of the generated data. 

We proceed in \textbf{4} main steps (Figure \ref{fig:Pipeline}), which are applied to all examples in FusedChat that feature prepended chitchat exchanges. 

\begin{figure}[bt]
\centering
\includegraphics[width=7.5cm]{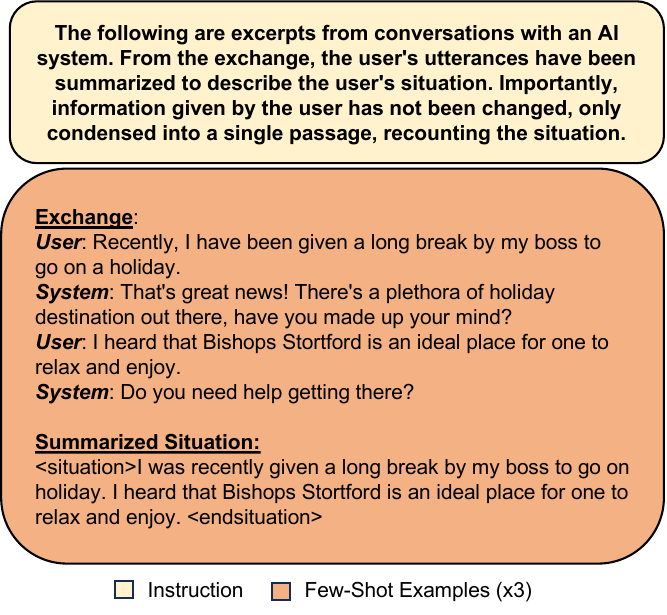}
\caption{\begin{small}Few-shot prompt (3 examples) for generating the seed situation. Prompts for generating the backstory and the reaction follow a similar structure.\end{small}}
\label{fig:FS_template}
\end{figure}

\paragraph{Seed Situation} 
We extract the prepended chitchat exchanges from FusedChat and generate summaries based on the information shared by the user. This summary serves as a seed situation that frames the interaction.  We create a prompt made up of an instruction prefix and three exemplars (Figure \ref{fig:FS_template}).

\paragraph{Turn Selection}
To increase the unpredictability of the interferences, we select a random exchange from the corresponding TOD continuation. In cases where the dialogue spans multiple domains (e.g., booking a train ticket and then a hotel), we ensure that the exchange is extracted from the first sub-dialogue, which is the most related to the seed situation. We achieve this by keeping track of the dialogue acts, which carry domain information. The cutoff point is determined by the turn that introduces an act from a different domain.

\paragraph{Exchange Augmentation}
The structure we consider aligns with the sequences presented in Figure \ref{fig:Interference}: elements of backstory are added \textit{after} the user's original utterance and a chitchat reaction is added  \textit{before}  the original system response. These components are generated in two steps. Each prompt provides an instruction and three exemplars that show the intended result. Moreover, each prompt example is structured with explicit separators to better guide the model's output. 
\begin{itemize}
     \item \textbf{Generating the user’s backstory.} We pass as input to the LLM a seed situation, any preceding conversation turns, and the user's task-oriented utterance.  Our prompt comprises examples that pair these inputs with a backstory which naturally follows the user's utterance (Appendix \ref{appendix:user_backstory}).

     \item \textbf{Generating the system's reaction.}  Given the previous dialogue turns, the original system response, and a newly-generated backstory, we prompt the LLM again to generate a chitchat reaction. In the prompt examples, we design the chitchat reactions to stay focused on the backstory, providing support and understanding (Appendix \ref{appendix:system_reaction}). While this type of chitchat may be limited in terms of diversity, it serves the purpose of avoiding new open-domain topics which could further derail the task interaction. This, in turn, helps the system maintain efficiency in task resolution, an essential attribute for TOD systems \cite{deriu2021survey}. 
\end{itemize}

\begin{table*}[t]
    \centering
    \begin{adjustbox}{width=16cm}
    \begin{tabular}{cccc}
    \toprule
         &  Total unique tokens&  Unique tokens not in MWOZ&  Avg. num. tokens per turn\\
         \midrule
         MWOZ &  18507 &  — &  13.13\\
         All Augmentated Turns &  25759 &  —&  28.38\\
         Turns w/ Backststory &  9419 &  6061&  36.7\\
         
        Turns w/ Reaction & 4998& 2770& 20.06\\
        \bottomrule
    \end{tabular}
    \end{adjustbox}
    \caption{\begin{small}Dataset statistics\end{small}}
    \label{tab: stats}
\end{table*}

\paragraph{Automatic filtering}
Similarly to \citealp{sun-etal-2021-adding}, we additionally automatically filter out backstories and reactions that are unlikely to be of good quality. The filters reject: (i) outputs that fail to adhere to the explicit prompt structure and lack the explicit separators, (ii) outputs which contain requestable slots from MultiWOZ (phone numbers, addresses…) as these are likely to repeat information from the original task utterance (iii) outputs that exhibit similarity to the utterance being augmented or that exactly contain the utterance being augmented. We use a Levenshtein similarity ratio of 50\% as a cut-off, which we found to work well in practice. \\

The augmented dataset contains TODs from MultiWOZ2.2 \cite{zang-etal-2020-MultiWOZ}, augmented with our desired scenario. The dataset includes a total of \textbf{3529} training examples, \textbf{458} validation examples and \textbf{488} test examples, after filtering. Relevant statistics can be found in Table  \ref{tab: stats}.  The interference-augmented user turns are notably longer than the average TOD turn, reflecting the conversational style of a chatty user who shares additional details about their request. In contrast, the reactions are comparatively shorter and less diverse, as their primary purpose is to provide support and understanding rather than introduce new information.

 \begin{table}[b]
     \centering
     \begin{tabular}{ccccc}
     \toprule
          &  Not at all &  Somewhat&  Fully& $\kappa$\\
          \toprule
          Q1&  0\% &  1.6\% &  98.4\% & 0.421\\
          Q2&  0.8\% &  12.3\% &  86.9\% & 0.236\\
          Q3&  6.6\% &  12.3\% &  81.1\% & 0.341\\
          \bottomrule
     \end{tabular}
     \caption{\begin{small}Averages for each rating on each question in \%. We also calculate Fleiss's $\kappa$ for each question to estimate inter-rater agreement.\end{small}}
     \label{tab:interference_eval}
 \end{table}

\paragraph{Human Evaluation}\label{para:human}
To control the quality of our examples, we perform an in-house human annotation of 25\% of the test set (a sample of 122 exchanges picked at random). We bring on 3 annotators who all have a background in linguistics, enabling them to potentially detect subtle inconsistencies in the exchanges. For each example, participants are presented with the generated user situation, the previous turns leading up to the augmented exchange and the exchange to be rated. Raters assess the exchange along 3 dimensions:
\begin{itemize}
    \item \textit{Q1. In the user turn, is the backstory being presented reasonable given the situation?}  This question assesses whether the model has followed instructions and based the user's backstory on the situation.  If the model generates content not explicitly present in the seed situation, raters are instructed not to penalize the output as long as it remains reasonable and coherent. For example, if the LLM adds \textit{I'm excited to go back!} to a backstory where the user is returning home to visit their family, the output should receive full marks.
    \item \textit{Q2. In the system's response, is the reaction provided supportive and understanding of the user's backstory ?} This dimension evaluates whether the LLM adheres to instructions and generates an appropriate reaction, consistent with our design choice.
    \item \textit{Q3. Overall, does the exchange sound natural and coherent ?} Finally we ask annotators to rate the exchange as a whole, to ensure  it retains its original coherence post-augmentation.
\end{itemize}
For each dimension, raters must choose between the following labels: \textit{Not at all, Somewhat,} and \textit{Fully}.
 
Examining Table \ref{tab:interference_eval}, human annotations reveal that the LLM effectively follows the given prompts and creates coherent exchanges overall, with fair to moderate agreement between annotators (Fleiss's $\kappa$). An average of only 6.6\% examples are rated as incoherent. Upon inspection of the selected exchanges, this rating arises when (i) the backstory amplifies an incoherence present in the original exchange (e.g. the original task response provides incorrect train times while the backstory emphasizes the importance of punctuality), (ii) when the randomly selected exchange makes it challenging to seamlessly incorporate the backstory (e.g. the backstory involves booking a museum in London while the selected exchange focuses on finding a museum in Cambridge) and (iii) the system reaction lacks commonsense (e.g. the reaction suggests that the airport is the user's final holiday destination). We treat these examples as noise to be expected in the dataset.

Overall, this readily implementable approach yields acceptable and natural exchanges as rated by humans, and lays the groundwork for future augmentation endeavors with LLMs for combining chitchat and TOD.

\section{Methodology and Experimental Setup} \label{sec:method}
\subsection{Methodology}
We choose the SimpleToD method \cite{hosseini-asl_simple_2020} to implement our baselines. Although the approach is typically trained with GPT-2 \cite{Radford2019gpt2}, we replace it with the more recent and state-of-the-art Llama-2-7B\footnote{\url{https://huggingface.co/meta-llama/Llama-2-7b}} \cite{touvron2023llama}, supplemented with LoRA \cite{hu2021lora}. This parameter-efficient technique has been shown to produce results comparable to full fine-tuning \cite{hu2021lora} and is therefore state-of-the-art for fine-tuning LLMs. It also produces a more regularized model, since a large portion of the pre-trained weights are left intact. This helps when facing unseen challenges during inference and makes this a powerful and effective method. We note that, as of yet, in-context learning approaches have not proven as effective for end-to-end TOD \cite{hudecek-dusek-2023-large, si2024spokenwoz}. We therefore adhere to the practice of fine-tuning in this work.

The SimpleToD method is a recent and popular end-to-end approach which relies on a single language model. We choose this method as it has been employed in prior research that explores chitchat-augmented TODs \cite{chen-etal-2022-ketod, sun-etal-2021-adding} and because it follows a very similar procedure to other strong baselines \cite{ham-etal-2020-end, yang2021ubar}. 

\begin{figure}[t]
\centering
\includegraphics[width=7.5cm]{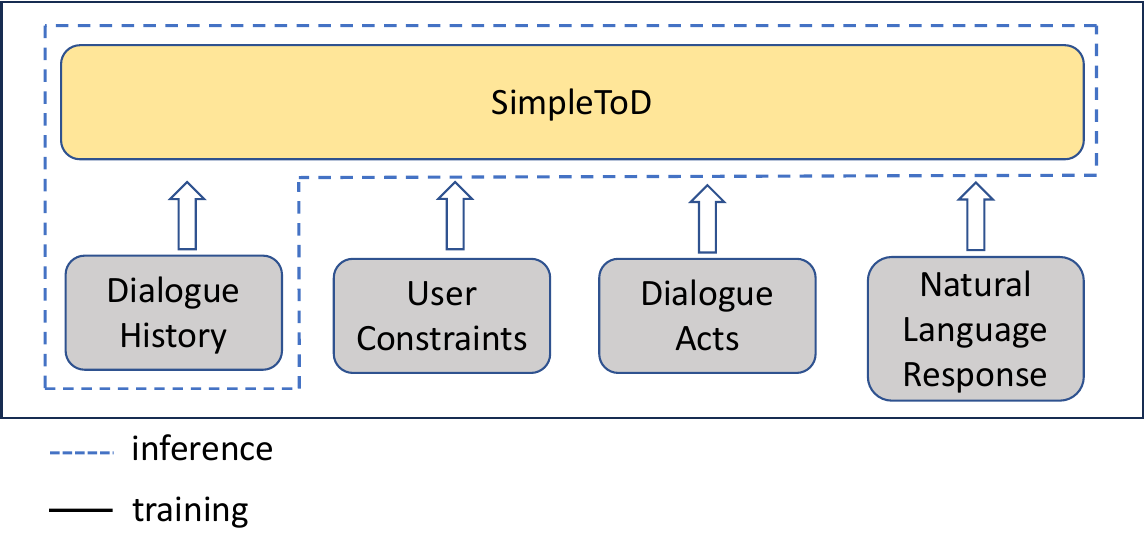}
\caption{\begin{small}An overview of the SimpleToD approach. At training time, a sequence of components is fed into a generative language model. During inference, only the dialogue history is used as input, and each component is generated in step-by-step autoregressive fashion.\end{small}}
\label{fig:SimpleToD}
\end{figure}

The SimpleToD method only takes as input the previous dialogue turns to sequentially generate three task-oriented sub-components in the form of text (Figure \ref{fig:SimpleToD}).  
\begin{itemize}
    \item First, the user’s constraints are generated in the form of [domain, slot, slot\_value] triplets. One possible example is [restaurant, food, indian].
    \item Subsequently, a collection of dialogue acts is produced, in the form of [domain, act, slot]. One possible example is [restaurant, inform, name]. 
    \item Finally, the model generates a response for the user (\textit{The name of the restaurant is The Golden Curry}). To ensure system adaptability to various databases, the model is often trained on \textit{delexicalized} responses, where entities to be fetched from the database are replaced by placeholders (\textit{The name of the restaurant is [name]}). This also simplifies the evaluation process when verifying the type of entity being offered by the system. 
\end{itemize}

The language model is fine-tuned on instances where [\textit{Dialogue History, User Constraints, Dialogue Acts, Response}] are concatenated into a single text sequence using special tokens as separators. During inference, we only pass [\textit{Dialogue History}] as input and the model is expected to generate the rest of the sequence. 

\subsection{Experimental Setup}
We propose two baseline variants and a model trained on our own augmented data, using the SimpleToD approach. The conditions for training and testing each variant are detailed in Table \ref{tab: train_test_conditions}. Distinctions between models are based on training data augmentations, as identical dialogue IDs are used in each set..

\begin{itemize}
    \item \textbf{SimpleToD} is a baseline representative of a typical task-oriented system: it is trained as intended by the SimpleToD authors. For reference, \textbf{SimpleToD-ref} is trained in the same way, but \textit{tested} on MultiWOZ examples.
    
    \item \textbf{SimpleToD-fused} is a baseline representative of a more versatile model, exposed to both TOD and chitchat turns during training. This model is trained on FusedChat examples which include prepended chitchat exchanges. Following the FusedChat authors' two-in-one approach, this model is trained to generate a task response when provided with a task input, and a chitchat response when given a chitchat input..

    \item \textbf{SimpleToD-inter} is trained on our own augmented examples. We use this model to verify that a model trained on our synthetic data can indeed 1) learn the MultiWOZ tasks adequately and 2) accommodate user chitchat with support and understanding while advancing the task in the same turn. No dataset has enabled models to do so thus far. 
\end{itemize}

\begin{table}[!ht]
\centering
\begin{adjustbox}{width=6.5cm}
\begin{tabular}{lcc}
\toprule
 Model          & Train         & Test \\
 \midrule
SimpleToD-ref   & MWoz      & MWoz \\
SimpleToD       & MWoz     & Interfere \\
SimpleToD-fused & FChat     & Interfere  \\
SimpleToD-inter & Interfere & Interfere  \\
\bottomrule
\end{tabular}
\end{adjustbox}
\caption{\small The same core MultiWOZ dialogues are used to train all models. \textit{MWoz} denotes the original, unaltered MultiWOZ dialogues; \textit{FChat} signifies dialogues are augmented as in FusedChat; and \textit{Interfere} indicates dialogues are enhanced with our custom augmentation. Distinctions between models are based on training data augmentations. Identical dialogue IDs are used across comparisons.}
\label{tab: train_test_conditions}
\end{table}

\begin{table*}[!ht]
\centering
\begin{adjustbox}{width=16cm}
\begin{tabular}{l|l|l|l|l|c|c|c|c}
\toprule
 & inform & success & JGA & CBE & unique tris. & BLEU-aug & BLEU-orig & BLEU-all \\
 \midrule
SimpleToD-ref & 80.73 &	72.85 & 0.66 & 1.87 & 2870 & – & – & – \\
\midrule
SimpleToD & 80.47 & 72.85 & \textbf{0.66} & 1.91 & 3184 & 0.11 & \textbf{0.24\textsuperscript{*}} & 0.21 \\
SimpleToD-fused & 67.08 & 51.38 & 0.39 & 2.12 & 5674 & 0.04 & 0.2 & 0.17 \\
SimpleToD-inter & \textbf{82.07} & \textbf{73.2} & 0.64 & \textbf{2.21} & \textbf{7561\textsuperscript{*}} & \textbf{0.25\textsuperscript{*}} & 0.22 & \textbf{0.22} \\
\bottomrule
\end{tabular}
\end{adjustbox}
\caption{\begin{small}Automatic evaluation results (baselines described in Sec. \ref{sec:method}). Values in bold indicate the best value for each metric and * indicates a statistically significant difference ($p<0.05$, paired t-test) between the best and the second best values in each respective column.\end{small}}
\label{tab: auto_eval}
\end{table*}

\paragraph{Implementation}
We train all models with the huggingface\footnote{\url{https://huggingface.co}} and pytorch \cite{paszke2017automatic} frameworks. For each variant, we perform a grid search over learning rates ranging from 1e-5 to 1e-4 with an effective batch size of 32 (batches of 16 with 2 gradient accumulation steps). We use learning rates of 3e-5 for all variants and train for 4 epochs. When performing LoRA fine-tuning, we choose the following LoRA configuration: a rank of size 32, given the complexity of our task, a scaling factor of 32, and target modules which include query and value projection matrices, found in the self-attention module of each transformer block.  Training and inference is done on a single 80Gb A100 GPU using 4 random seeds for each experiment. After training, we generate outputs using greedy decoding as in SimpleToD and prevent the model from repeating n-grams of size 10, to avoid repetitive loops that may arise.

\paragraph{Automatic Evaluation Metrics}
We employ the standard evaluation metrics used for MultiWOZ\footnote{\url{https://github.com/Tomiinek/MultiWOZ_Evaluation?tab=readme-ov-file}} \cite{nekvinda-dusek-2021-shades}. The \textbf{joint goal accuracy} (JGA) reflects the proportion of turns where the predicted user constraints exactly match the gold ones. The \textbf{inform rate} evaluates the system’s capacity to provide the right type of entities from the database, given the user's constraints, and the \textbf{success rate} assesses how effectively the system delivers requested attributes like phone numbers or booking references. For a more comprehensive understanding of these metrics, we direct readers to the MultiWOZ paper \cite{budzianowski-etal-2018-MultiWOZ}. 
For response quality, we use three BLEU \cite{papineni-etal-2002-bleu} scores as in \cite{chen-etal-2022-ketod}: \textbf{BLEU-aug} rates responses that follow interferences, \textbf{BLEU-orig} evaluates all other responses and \textbf{BLEU-all} evaluates all responses holistically.
For response diversity, we analyze the \textbf{Conditional Bigram Entropy} (CBE) and the count of \textbf{unique trigrams} to gauge the richness of vocabulary and phrasing in the generated responses \cite{nekvinda-dusek-2021-shades}.

\section{Results and Discussion} 
\label{sec: results}

\paragraph{Automatic Evaluation Results}
The results of the automatic evaluation in Table \ref{tab: auto_eval} suggest that our chosen \textbf{SimpleTOD baseline} is quite robust to user backstories. Despite the interferences injected in the test set, SimpleToD perfoms as well as SimpleToD-ref, staying on track with regards to the task.  We expect this baseline's robustness to be related to our choices of model and fine-tuning technique. Llama-2 and LoRA's regularization-friendly setup aids in maintaining task focus when confronted with novel scenarios like the interferences we introduce.  However, BLEU-aug shows that responses to user turns with chitchat interferences are not satisfactory, as we further show in our human evaluation below.

On the other hand, \textbf{SimpleToD-fused} encounters challenges when presented with the interference-augmented test set. This model is specifically trained to respond to either a chitchat turn or a task-oriented turn, and as a result, falls into one mode or the other when confronted with our interferences.  In most cases, it leans towards continuing with chitchat, resulting in lower inform and success rates. This also results in very low BLEU-aug scores, primarily because the TOD segment is frequently absent from the response, and also because the chitchat that is generated does not closely match the expected reactions. This reveals that training on dialogues where chitchat and TOD are depicted as mututally exclusive response modes falls short in addressing inter-mode user turns.   

The \textbf{SimpleToD-inter} model, trained on our augmented data, delivers task-oriented performance comparable to that of SimpleToD, with no statistically significant differences observed. Moreover, it aligns as expected with the augmented system responses, as indicated by the Bleu-aug score. Consequently, this shows that our generated training data enhances the model's adaptability to potentially chatty users, without negatively impacting its task performance.

\begin{table}[t]
    \centering
    \resizebox{7.5cm}{!}{
    \begin{tabular}{ccccc}
    \toprule
         &  \#1 &  \#2&   \#3 & Mean Rank\\
         \midrule
         SimpleToD& 11.33\%  & \textbf{77.33}\%  & 11.33\%  & 2.0  \\
         & \scriptsize{$\pm$ 10.37} & \scriptsize{$\pm$ 7.36} & \scriptsize{$\pm$ 4.99} & \scriptsize{$\pm$ 0.15} \\
         SimpleToD- & 5.33\%  & 33.33\%  & \textbf{61.33}\%  & 2.56 \\
         fused  & \scriptsize{$\pm$ 2.49} & \scriptsize{$\pm$ 10.5} & \scriptsize{$\pm$ 10.87} & \scriptsize{$\pm$ 0.12} \\
         
        SimpleToD- & \textbf{92.0}\%  & 6.67\%  & 1.33\%  & \textbf{1.09}  \\
        inter & \scriptsize{$\pm$ 3.27} & \scriptsize{$\pm$ 3.4} & \scriptsize{$\pm$ 0.94} & \scriptsize{$\pm$ 0.03} \\
 \bottomrule
    \end{tabular}
    }
    \caption{\begin{small}Human rankings of responses from each baseline, following a user request augmented with backstory. The percentages represent the distribution of rankings per model. Rank 1 is best, so a lower rank implies preferred responses.\end{small}}
    \label{tab:human_ranking}
\end{table}

\paragraph{Human Evaluation Results} \label{para:human_eval}
To better evaluate the output quality of our variants, we conduct an in-house human evaluation of delexicalized responses to user turns with interferences. Three evaluators, distinct from those used in our initial annotation, participate in this task. They all possess expertise in NLP, making delexicalized responses pose no issue. To create a simple and straightforward annotation, we adopt a methodology inspired by  \citealp{nekvinda-dusek-2022-aargh} which consists in side-by-side relative ranking. This approach has demonstrated enhanced consistency when contrasted with evaluating individual instances in isolation \cite{kiritchenko-mohammad-2017-best}. We randomly select 50 examples from the test set, singling out the turns that contain our introduced interferences.

For each example, participants are presented with the dialogue history up to the augmented user turn and three system responses to be ranked. The responses are displayed side by side, each of them having a dedicated scale from 1 to 3, indicating the quality from best to worst. Annotators are explicitly instructed to focus on how well each response helps the user with their booking \textit{and} on how naturally the response follows the previous user turn. If the evaluators deem it relevant, they can assign the same rating to multiple responses. 

Results in Table \ref{tab:human_ranking} show a persistent preference for SimpleToD-inter: this model's responses both move the task forward and follow more naturally after the user's turn. Regarding annotator agreement, we find that standard deviations offer more insight than Fleiss's $\kappa$, and note that there is  high agreement on ranking SimpleToD-inter responses as the top choice (\#1). However, the level of agreement is lower for the other two baselines, even though there is a clear inclination toward the responses from SimpleToD.

We inspect the rated responses, showing some illustrative examples in Table \ref{tab:response_examples}. To characterize each variant, we employ \citealp{taleb2014antifragile}'s Fragile-Robust-Antifragile framework, which has previously been applied in the context of Information Systems \cite{Gorgeon-antifragile}.  We  adapt this framework to assess how each variant handles inter-mode inputs that potentially disrupt the flow of a TOD.

Responses from SimpleToD-fused typically belong to the "Fragile" category. This model, when presented with a user backstory after a request, often produces responses focused more on chitchat, neglecting the original request. Such behavior slows down the task-oriented dialogue and increases the likelihood of the conversation deviating from its intended path.

\begin{table*}[!t]
\label{tab: resp_examples}
    \centering
    \resizebox{16cm}{!}{
    \begin{tabular}{p{4.0in}|p{4.0in}}
    \toprule
      \textbf{\textit{Dialogue History}}   &  \textbf{\textit{Responses}} \\
         \midrule
         \textit{User}: I need a taxi. I am going to Avalon and I need to leave after 16:15 & \textbf{SimpleToD}: I have booked a [type] for you. The contact number is [phone].\\
         \textit{System}: I will need to know where you need to be picked up? &  \textbf{SimpleToD-fused}: I hope you have a great time!\\
          \textit{User}: I will be at Parkside Pools. \hl{I'm so excited for later! My crush just invited me to her birthday party at her house, Avalon. I'm super nervous right now!} & \textbf{SimpleToD-inter}: Aww, that's so sweet! I'm sure you'll have a great time at the party. I have a [type] for you. The contact number is [phone].\\
        
        \midrule
        User:  I'd like to take a train from Cambridge on Sunday... 
        
        System: ...TR7447 will get you there by 13:51. Do you need some seats booked? 
        
        User: Yes. Can you please book 2 seats? \hl{I'm actually going to the dentist on Sunday. I have a toothache and I'm hoping to get it fixed. I'm trying to soothe the pain with ice until then...} 
        &
        
        \textbf{SimpleToD}:  Booking was successful, the total fee is [price] payable at the station. Reference number is: [ref]. Is there anything else I can help you with? 
         
        \textbf{SimpleToD-fused}: Booking was successful, the total fee is [price] payable at the station. Reference number is: [ref]. Is there anything else I can help you with? 
        
        \textbf{SimpleToD-inter}: Sorry to hear that you're in pain. I hope the dentist appointment goes well and that you get some relief from the toothache. I have booked [bookpeople] tickets for you. Your reference number is [ref] . Is there anything else I can help you with?\\ 

         \midrule
        User: I am departing from Cambridge on Friday. \hl{I'm actually going to visit my friend in the hospital. He got into a motorbike accident and I've been worried sick about him. I've always told him that motorbiking is dangerous, but he wouldn't listen. Now he's paying the price. I hope he's doing alright...} & \textbf{SimpleToD}: I'm sorry to hear about your friend. I'm sure he'll be ok. I have train [trainid] arriving at [arriveby] . Would you like me to book it for you? 
        
        \textbf{SimpleToD-fused}: I'm sorry to hear about your friend. I hope he's okay. 
        
        \textbf{SimpleToD-inter}: Sorry to hear about your friend. I hope he recovers quickly. [trainid] leaves at [leaveat] and arrives at [arriveby]. Would you like to book a seat?\\
         \bottomrule 
    \end{tabular}
    }
    \caption{\begin{small} Examples of system responses following augmented user turns (dialogues SNG0055, PMUL0994 and PMUL1424). These serve to demonstrate how SimpleToD-fused alternates between task-oriented and chitchat responses, with  chitchat being predominant. Example \#3 highlights SimpleToD's occasional capacity to offer support alongside addressing the task request. However, this behavior is not as consistent as in the case of SimpleToD-inter, the model favored by human raters and trained on our augmented dataset.\end{small}}
    \label{tab:response_examples}
\end{table*}

SimpleToD can be classified as relatively "Robust." It predominantly disregards the elements of backstory in the user turn and maintains a strong focus on accomplishing the task. However, this approach results in responses that may appear detached and dismissive of the user, leaving a negative impression that could dissuade users from returning to the system. Interestingly, in some rare instances, the model demonstrates the capacity to generate supportive reactions such as \textit{I'm sorry to hear about...} before reorienting its answer back to the task, as seen in example \#3, Table \ref{tab:response_examples}. This observation can be attributed to our model and fine-tuning choices.

To an extent, SimpleToD-inter demonstrates characteristics that align with being "Antifragile," a term which refers to being able to benefit from potential disturbances. In this context, inter-mode inputs are welcomed and handled accordingly, enhancing the overall interaction compared to more typical TOD flows. This creates a more engaging dialogue and ultimately incentivizes the user to continue using the system. 

Nonetheless, in terms of limitations, we note that responses exhibit somewhat limited diversity, which could lead to users anticipating the system's reactions. Moreover, while we have addressed one specific scenario, the challenge of making a single system antifragile to a variety of scenarios remains.  However, given our encouraging results, generating inter-mode scenarios automatically and then training a TOD system in a multi-task fashion holds promise.

\section{Related Work}
The integration of chitchat into task-oriented dialogues is a growing trend in recent research. For instance, Accentor \cite{sun-etal-2021-adding} adds automatically generated, human-filtered chitchat to system responses. However, their chitchat often lacks specific contextual relevance to the ongoing conversation, resulting in responses like \textit{Sounds good!}. On the other hand, KETOD \cite{chen-etal-2022-ketod} expands system responses by introducing knowledge-grounded chitchat. This chitchat relies on specific, contextually relevant information, leading to more diverse enhancements compared to Accentor. While the knowledge is automatically extracted from Wikipedia, system responses are refined by human annotators. Although valuable, both of these approaches have limitations as they exclusively augment system turns, assuming the user input will stay fixed.

Other work takes an interest in adding full chitchat exchanges to TODs. For example, FusedChat \cite{young2022fusing} prepends and appends human-generated chitchat. Recently, \citeauthor{liu-etal-2023-system-initiated} build on this dataset to explore system-initiated transitions from chitchat to task-oriented dialogue. Additionally, \citeauthor{li-etal-2023-enhancing-task} automatically inject chitchat exchanges within MultiWOZ dialogues, creating more engaging and flexible TODs. While these contributions are valuable, they typically treat individual turns as exclusively task-oriented or chitchat-oriented, without allowing for potential overlap. Our work represents a step towards addressing more ambiguous user turns that may contain elements of both modes simultaneously.

\section{Conclusion}
We portray chitchat in TODs as a natural interference, introducing a novel task in which users incorporate elements of backstory into their requests. We generate this scenario in an automated and controlled manner by employing few-shot prompting with a large language model. This method produces exchanges human evaluators deem acceptable and natural. Our study reveals that a commonly used TOD baseline cannot fully handle this setting and neither can its variant trained on a mixture of TOD and chitchat turns. In contrast, a model trained on our enriched dataset consistently and seamlessly handles the introduced chitchat while effectively advancing the task at hand.  We believe our work is an important step in automating the creation of diverse and challenging chitchat scenarios within TODs, ultimately leading to the emergence of more resilient TOD systems.

\section*{Ethics Statement}
Our dataset is the product of automated generation, which means it could potentially contain biases from the data used to train the underlying LLM. Nervertheless, this LLM is open-source, making it possible for the wider research community to thoroughly examine these biases. Furthermore, it is important to note that our training data is not guaranteed to exclusively contain supportive responses. Therefore, it should be employed judiciously, being better suited for low-stakes settings or research purposes.

\section{Acknowledgments}
This work was granted access to the HPC resources of IDRIS under the allocation 20XX-AD011014510 made by GENCI.

\section{Bibliographical References}\label{sec:reference}
\bibliographystyle{lrec-coling2024-natbib}
\bibliography{Chitchat}

\appendix

\section{Prepended Chitchat in FusedChat} \label{appendix:fusedchat_example}
Refer to Table \ref{tab:Fchat_example}.

\begin{table}[b]
\centering
    
    \begin{tabular}{p{2cm}p{4.8cm}}
        \toprule
        User (\textit{CC})& I am meeting my client in Cambridge soon. I'm kind of nervous. \\
        \midrule
        System (\textit{CC}) & Is it an important meeting?  \\
        \midrule
        User (\textit{CC}) & This is my first client.  \\
        \midrule
        System (\textit{CC}) & Wow that is huge! Good luck! \\
        \midrule
        User (\textit{Task}) & I am looking for a train that will be arriving there by 16:00 Friday, from King's Lynn. \\
        \midrule
        System (\textit{Task}) & We have a train headed for Cambridge at 15:11. Would you like to book it? \\
        \midrule
        ...& ...\\
        \bottomrule
       \end{tabular}

       \caption{ \small A full chitchat exchange, related to the task, is prepended by human annotators to a MultiWOZ dialogue. \textit{CC} indicates chitchat turns and \textit{Task} indicates task-oriented turns. The ellipses indicate the continuation of the original MultiWOZ dialogue. }
    \label{tab:Fchat_example}
\end{table}

\section{User Backstory Prompt}\label{appendix:user_backstory}
Refer to Table \ref{tab: backstory_prompt}.

\begin{table}[h]
\centering
    \begin{tabular}{p{7.2cm}}
        \toprule
        In the following examples, you are presented with a user's situation and a conversational context, which may be None if it is the start of the conversation. The user shares their backstory by adding it to their original utterance. Their backstory is based on the user's situation and should naturally follow the original utterance. It should be very fluent and coherent with the conversational context.\\
        
        \midrule
        \midrule
        
        Situation: 
        
        I was talking to my brother on the phone earlier today. He's getting married We discussed his wedding plans and decided to meet up at the London Liverpool Street train station today. \\
        
        \midrule
        Conversational Context:  
        
        \textbf{User}: I would like for a taxi to take me to london liverpool street train station, arriving no later than 17:45 please.
        
        \textbf{System}: I can book that for you, first I'll need to know where you'd like picked up at. \\
        
        \midrule
        Original User Utterance: 
        
        \textbf{User}: **I would like to depart from London Kings Cross Train Station.** \\
        
        \midrule
        User Utterance With Backstory: 
        
        \textbf{User}: **I would like to depart from London Kings Cross Train Station.** + <Backstory: My brother is getting married! I was talking to him on the phone earlier and we decided to meet at the London Liverpool train station.> [END]\\
        
    \bottomrule
       \end{tabular}

       \caption{\small An example prompt for generating backstories. The first row contains the instruction prefix, added at the start of the prompt. All subsequent rows represent the components of the prompt, added in that order, to create a single exemplar.  Our prompt contains 3 such exemplars.}
    \label{tab: backstory_prompt}
\end{table}

\section{System Reaction Prompt}\label{appendix:system_reaction}
Refer to Table \ref{tab:reaction_prompt}.

\begin{table}[h]
\centering
    \begin{tabular}{p{7.2cm}}
        \toprule
        In the following examples, you are presented with a conversational context. In the last turn, the user shares their backstory. The original system response should be improved to include a reaction to the user's backstory at the beginning of the response. This reaction should be supportive and display an understanding of the user's situation. It should be unique to the backstory and contextual to the conversational context. Avoid repeating expressions found in previous examples.\\
        
        \midrule
        \midrule
        Conversational Context:  
        
        \textbf{User}: I would like for a taxi to take me to london liverpool street train station, arriving no later than 17:45 please
        
        \textbf{System}: I can book that for you, first I'll need to know where you'd like picked up at.
    
        \textbf{User}: I would like to depart from London Kings Cross Train Station. <Backstory: My brother is getting married! I was talking to him on the phone earlier and we decided to meet at the London Liverpool train station.>\\
        
        \midrule
        Original System Response: 
        
        \textbf{System}: **A white Toyota is booked for you.** \\
        
        \midrule
        Response With Reaction: 
        
        \textbf{System}: <Reaction: I see! Congratulations to your brother!> + **A white Toyota is booked for you.** [END]\\
        
    \bottomrule
       \end{tabular}

       \caption{\small An example prompt for generating chitchat reactions. The first row contains the instruction prefix, added at the start of the prompt. All subsequent rows represent the components of the prompt, added in that order, to create a single exemplar.  Our prompt contains 3 such exemplars.}
    \label{tab:reaction_prompt}
\end{table}

\end{document}